\documentclass[conference]{IEEEtran}
\IEEEoverridecommandlockouts

\usepackage[utf8]{inputenc} 
\usepackage[T1]{fontenc}    
\usepackage[draft]{hyperref}       
\usepackage{url}            
\usepackage{booktabs}       
\usepackage{amsfonts}       
\usepackage{nicefrac}       
\usepackage{microtype}      

\usepackage[pdftex]{graphicx}

\usepackage[square,numbers]{natbib}

\usepackage[T1]{fontenc}
\usepackage{lmodern}
\usepackage[left=2cm, right=2cm, top=2cm]{geometry}

\title{Multi-Agent Deep Reinforcement Learning with Human Strategies}
\date{\vspace{-3ex}}

\author{
  Thanh Nguyen, Ngoc Duy Nguyen, and Saeid Nahavandi\\
  Institute for Intelligent Systems Research and Innovation (IISRI)\\
  Deakin University, Waurn Ponds Campus\\
  Geelong, VIC, 3216, Australia\\
  \texttt{E-mails:\{thanh.nguyen,duy.nguyen,saeid.nahavandi\}@deakin.edu.au} \\
}

\begin{document}

\maketitle

\begin{abstract}
Deep learning has enabled traditional reinforcement learning methods to deal with high-dimensional problems. However, one of the disadvantages of deep reinforcement learning methods is the limited exploration capacity of learning agents. In this paper, we introduce an approach that integrates human strategies to increase the exploration capacity of multiple deep reinforcement learning agents. We also report the development of our own multi-agent environment called Multiple Tank Defence to simulate the proposed approach. The results show the significant performance improvement of multiple agents that have learned cooperatively with human strategies. This implies that there is a critical need for human intellect teamed with machines to solve complex problems. In addition, the success of this simulation indicates that our multi-agent environment can be used as a testbed platform to develop and validate other multi-agent control algorithms. 

\end{abstract}

\section{Introduction}
\label{sec:1}
Reinforcement learning (RL) \cite{1} has attracted wide research interests due to its ability to plan a long-term solution for sequential decision making. However, traditional RL algorithms seem to work in obvious problems where the number of data dimensions is limited. To deal with multi-dimensional environments, a deep neural network is used to approximate the probability distribution of all possible actions \cite{27,30}. However, the use of neural networks may cause divergence while estimating Q-value function \cite{7}. 

The primary reason of this chaos originates from RL's online learning, i.e., sequential data from the environment contain correlative constraints. Therefore, every update of the Q-value function tends to approximate a biased value and forces the agent into local minimum solutions. As a result, a number of algorithms have been proposed in the literature to handle the correlative information of input data. For instance, correlative samples are first assembled and stored in an \emph{experience replay} memory \cite{2}, and are later selected randomly to estimate the Q-value function. Using different network (known as a \emph{target network} \cite{2}) also avoids correlative samples because the network is only updated infrequently \cite{28}. \emph{Double Q-learning} \cite{8} is another approach that performs action selection and action evaluation in two separate networks. To further enhance correlative dilation, each sample in the replay memory has a priority level depending on its \emph{temporal difference error} \cite{9}. Moreover, particularly in stochastic environments, an asynchronous method proves helpful, i.e., the goal is to create multiple, simultaneous agent-environment instances. In other words, agents are trained separately in different replica environments, but afterward they jointly update a shared policy network \cite{10}. Additionally, Open AI \cite{12} uses a heuristic search to optimize an objective by conducting evolutionary generations through thousands of workers.

\begin{figure}[!t]
\centering
\includegraphics[width=0.9\linewidth]{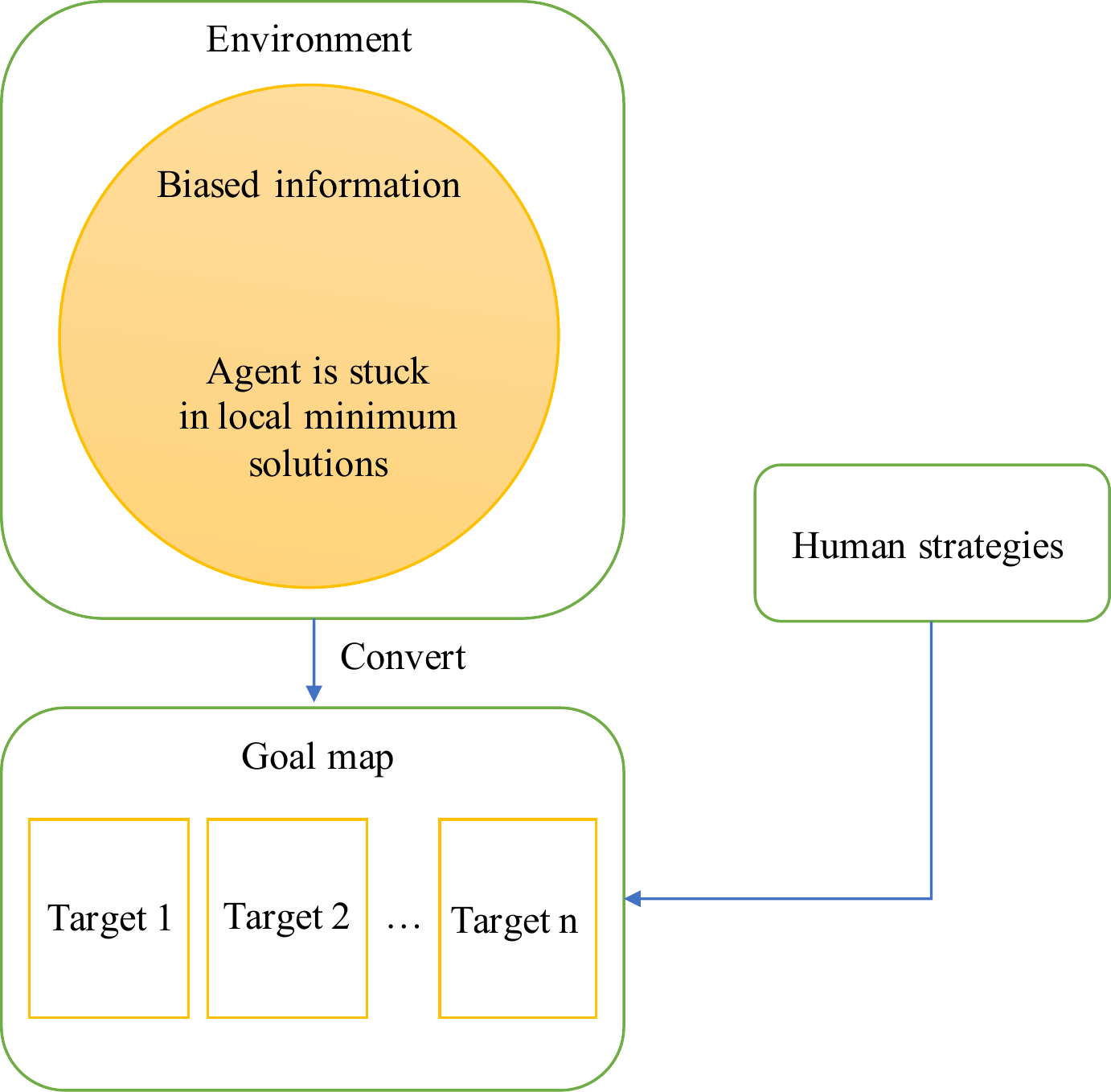}
\caption{Using a goal map to tackle a biased environment.}
\label{fig:1} 
\end{figure}

However, what if an observed environment is profoundly biased (e.g. the Multiple Tank Defence game described in Section \ref{sec:A}), or if the biased data take a primary portion of the environment's data distribution? In these cases, all the algorithms mentioned above may no longer work because the agent is tricked to follow a greedy solution. In the worst cases, the agent turns into a ``zombie'' and behaves peculiarly. One straightforward solution is to extend the training time or to allocate more computational power resources such that further exploration may be attained. However, extending the training time is not desirable, and may even prove intractable if the observed environment is fully biased. For instance, given an agent that controls a fighter jet over a battlefield with a variety of enemies, the jet will crash if it collides with an enemy or if it runs out of fuel. The jet may replenish its fuel by flying over a fuel depot. However, shooting the fuel depot will earn a high reward, which tricks the agent to shoot all the fuel depots to assure the highest score achievement. As a result, the agent rarely advances to the next stage because of fuel depletion and becomes stuck in a local minimum solution. Multiple Tank Defence is another example of such an environment and will be described in Section \ref{sec:A}.

In this paper, we tackle the biased environments by using human strategies. Initially, we use human strategies to represent the environment into a \emph{goal map}. In the goal map, we define a set of \emph{targets} which each presents a human-centric goal. For instance, in the previous example, the desired agent should control the jet to shoot fuel depots to accumulate rewards but should recognize when fuel is low, and refill the fuel. The goal map in this case includes two essential targets: one target represents for any obstacles that the jet can shoot to accumulate rewards and another target represents for the action of refilling the fuel. By using this approach, the player can prolong the jet's lifetime and achieve a higher score in the long term. The next step is to train the agent to understand the goal map and achieve the predefined targets in the goal map. Therefore, we develop a software architecture called \emph{Multi-Target System} (MTS) that is used to train agents to understand the goal map and enable cooperation between multiple agents. Fig.~\ref{fig:1} illustrates our proposed method to tackle a biased environment by using human strategies and a goal map. Section \ref{sec:3.4} describes MTS in more detail. 

In summary, the paper includes the following contributions:

\begin{itemize}
\item We developed an environment named Multiple Tank Defence that is used as a testbed to evaluate the performance of deep RL methods in multi-agent settings. Moreover, the use of Multiple Tank Defence is similar to \emph{Arcade Learning Environment} \cite{13}. As a result, we can use any deep RL methods to solve the Multiple Tank Defence without changing existing parameters.

\item The goal map can be utilized in different practical scenarios such as aiding the cooperation between multiple agents or controlling agent behaviors in running time without asking a lot of human feedback.

\end{itemize}

This paper includes the following sections. The next two sections summarize the related work and introduce the gameplay of Multiple Tank Defence. Section \ref{sec:3} introduces the goal map and its implementation by using MTS. Section \ref{sec:4} presents the performance evaluation of the proposed method in Multiple Tank Defence. Finally, Section \ref{sec:5} concludes our work.

\section{Related Work}
\label{sec:2}

\begin{figure*}[!t]
\centering
\includegraphics[width=0.7\linewidth]{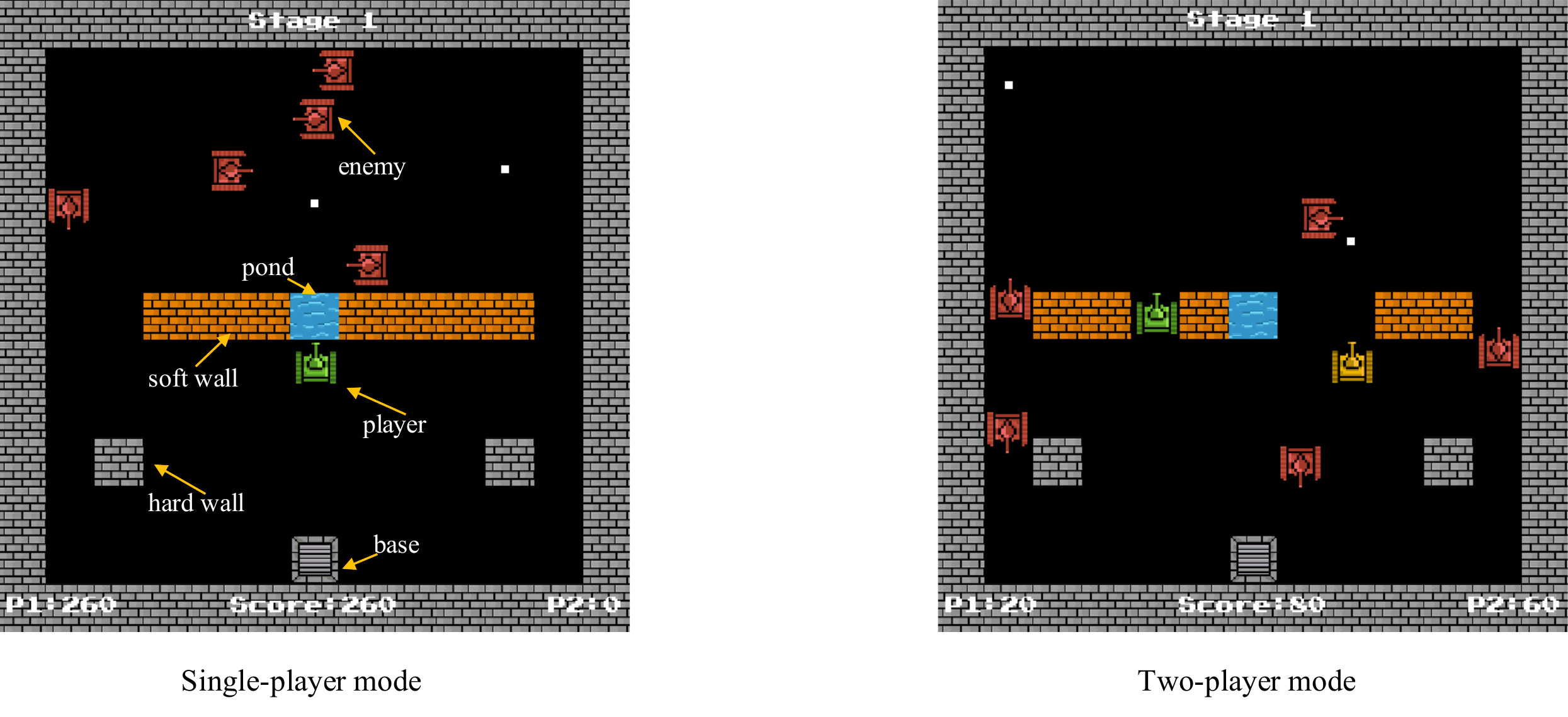}
\caption{A gameplay interface of Multiple Tank Defence.}
\label{fig:A1} 

\end{figure*}
\begin{figure*}
\centering
\includegraphics[width=1.0\linewidth]{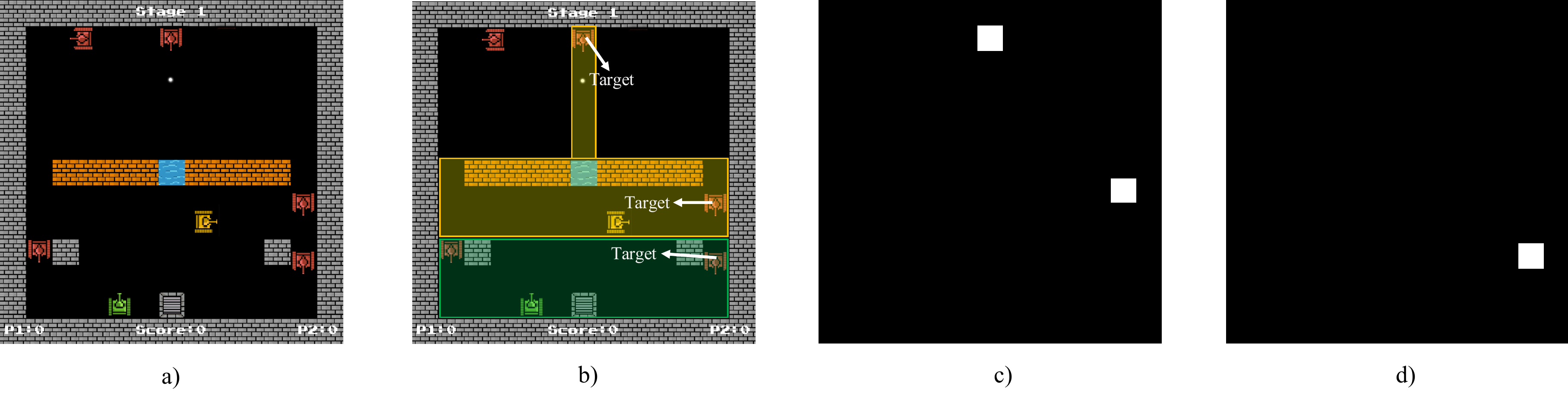}
\caption{An example of using a goal map in Multiple Tank Defence. a) A game screen in Multiple Tank Defence. b) A goal map (contains human strategies) in the game. c) A set of targets of yellow player is represented as a mask. d) A target of green player is represented as a mask.}
\label{fig:2} 
\end{figure*}

Since 2015, there have been a number of improvements over DQN \cite{2} such as \emph{double Q-network} \cite{8}, \emph{dueling network} \cite{14}, and \emph{prioritized experience replay} \cite{9}.
However, these extensions require a large amount of resources to train agents regarding training time and memory allocation. In 2016, Mnih \emph{et al.} \cite{10} introduced a light-weight approach based on actor-critic architecture \cite{15}, namely \emph{asynchronous advantage actor-critic} (A3C), by training multiple agent-environment instances at the same time. This method reduces training time, the use of memory allocation, and becomes a state-of-the-art method in Atari domain. As a result, we use A3C algorithm as the baseline in this paper. 

Alongside the rapid increase in the complexity of applications, there has been a rising demand to provide human feedback in the agent training process \cite{29}. Christiano \emph{et al.} \cite{17} reshape the reward signal by injecting human feedback during the training process. However, this approach requires an expert to observe hundreds to thousands of agent's video clips. Our proposed method is more feasible as we create a goal map, which integrates human strategies before the training process. Moreover, the goal map is easily created by using an essential localization method \cite{18}.

Another challenge in RL is navigation of environments with sparse rewards. In such cases, agents become easily stuck in local minimum solutions due to insufficient exploration. One solution is to split a complicated task into hierarchical subtasks in which a parent subtask has higher abstraction than its successors \cite{20}. By combining hierarchical subtasks with the agent's intrinsically-motivated rewards \cite{21}, Kulkarni \emph{et al.} \cite{22} successfully instructed an agent to accomplish its goal in the Atari game \emph{Montezuma's Revenge}. However, this approach requires training of two policies simultaneously: one that estimates goal-value function and one that estimates action-value function. By adopting temporal abstraction with the target map, our proposed scheme does not require to train an additional policy network over goals and enables smooth cooperation within agents and also between humans and agents. Moreover, as a target map is independent of policy networks and deep RL algorithms, our approach is more robust and can deal with a broader range of applications.

Finally, our proposed method is used to cooperate multiple agents in multi-agent systems. Previous studies \cite{23, 24} train agents to communicate in a shared limited bandwidth channel. However, these methods limit a number of training agents. Recently, Gupta \emph{et al.} \cite{25} scale to cooperate a huge number of agents by using a parameter-sharing network. This approach is used only in homogeneous systems. In this paper, we introduce a goal map that can be used to train agents with different types. Section \ref{sec:3.4} discusses this issue in greater detail.

\section{The Multiple Tank Defence Environment}
\label{sec:A}

In this section, we introduce the gameplay of Multiple Tank Defence, as illustrated in Fig.~\ref{fig:A1}. The game includes one or two player agents and five enemy tanks. If an enemy tank is destroyed, another enemy will appear randomly on the battlefield. This approach assures the stochastic behavior of the game. The game is over if the base is destroyed or no players left. Finally, a reward of 10 is given for destroying an enemy tank. 

There are different terrains in the game such as a pond tile, a soft wall tile, and a hard wall tile. A pond tile is used to avoid a tank from moving though a bullet can pass through it. A soft wall can be collapsed by a bullet, but a hard wall cannot be demolished. Different terrains can be constructed in different locations on the battlefield to enable a complicated strategy to protect the base. Multiple Tank Defence is a biased problem as the agents greedily destroy the enemies to accumulate rewards without protecting the base. As the result, the game is over earlier.

Moreover, Multiple Tank Defence includes the following features:

\begin{itemize}

\item Multiple Tank Defence can have unlimited number of stages by modifying the number of enemies, the number of players, and different construction ways of terrains on the battlefield. 

\item The game has two different goals: one goal is to destroy enemies to accumulate the rewards and another goal is to protect the base to prolong the game lifetime. Therefore, the game requires a complicated strategy to get a high score.

\item The game supports human mode, i.e., a human can play with an AI agent to cooperatively protect the base. This requires the agent to follow a designated strategy.

\end{itemize}

\section{Proposed Scheme}
\label{sec:3}

In this section, we introduce the concept of a goal map. We then describe the implementation detail of a goal map by using masks. Finally, we propose a network architecture that is used to train agents to understand the goal map.

\subsection{Goal map}
\label{sec:3.1}

A goal map is an abstract graphical representation that is used to present the human strategies while analyzing the problem. Fig.~\ref{fig:2}b illustrates an example of a goal map in Multiple Tank Defence. In the goal map, we define a set of targets for each agent. Specifically, the yellow agent's targets are all enemies inside the yellow region and the green agent's target is the enemy that is closest to the base. 

\subsection{Implementation details}
\label{sec:3.2}

To efficiently implement the goal map, we introduce the concept of a goal mask for each agent. The goal mask is a black image in which contains a set of targets. A target is then represented by a white square, as shown in Fig.~\ref{fig:2}c and Fig.~\ref{fig:2}d.

Specifically, the goal masks have the same dimension with the observed graphical state of the environment. The mask is then converted into grayscale and it is rescaled by a factor $\delta < 1$. Other information of the targets such as working region, priority target, reward information, or target location is stored in a metadata structure. Those metadata are retrieved later in the training process to instruct agents to achieve the targets defined in the goal map.

\begin{figure}[!h]
\centering
\includegraphics[width=1.0\linewidth]{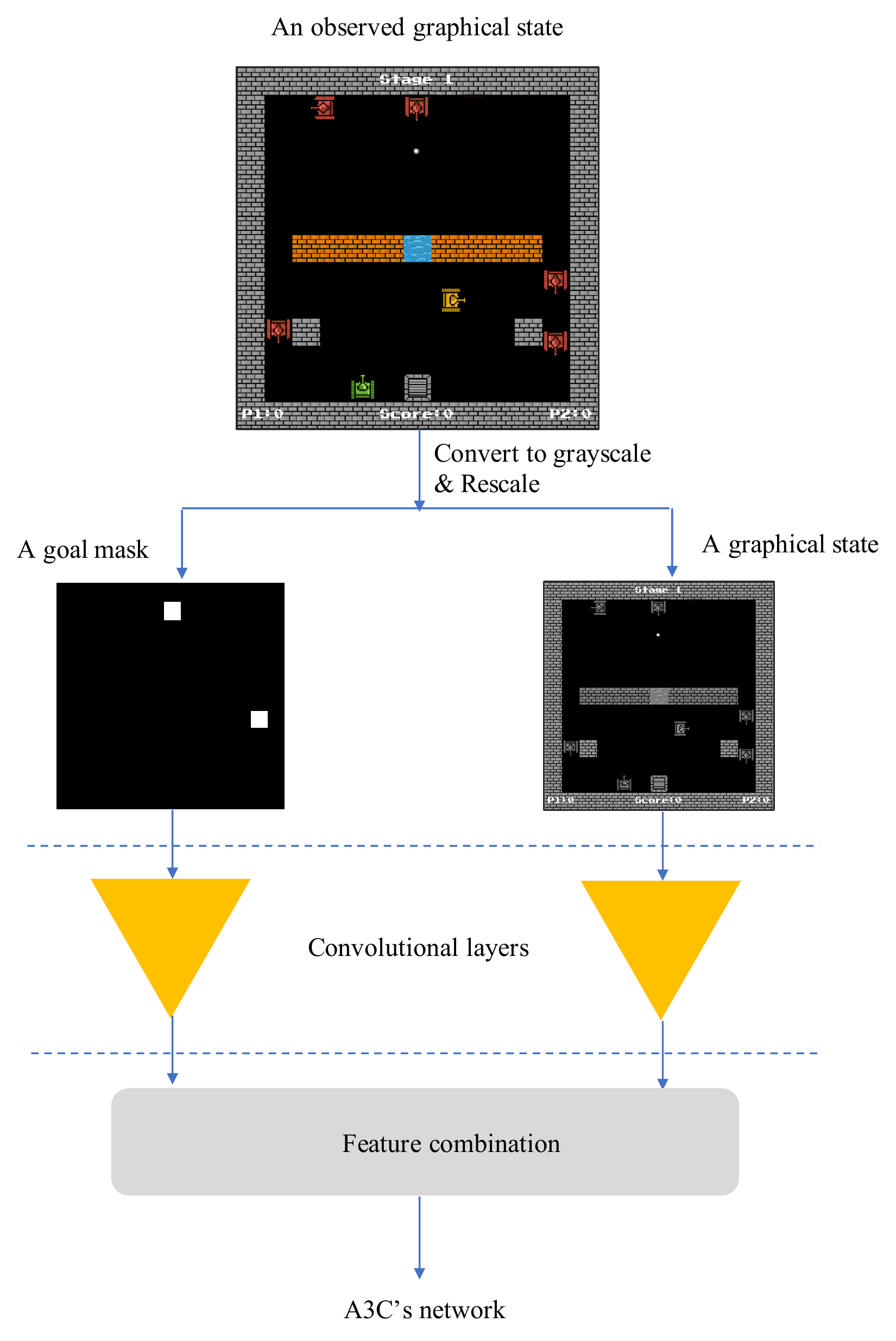}
\caption{A policy network for the yellow tank in Multiple Tank Defence.}
\label{fig:3} 
\end{figure}

\subsection{Training process}
\label{sec:3.3}

In the training process, the key point is to create a separate policy network for each agent or a set of agents, which have the same strategy. Specifically, we create two policy networks in Multiple Tank Defence to represent the goal map described in the previous subsection. Fig.~\ref{fig:3} illustrates a policy network for the yellow region. Initially, we generate the goal mask and the grayscale version of the original state. Those data are put into two separate convolutional networks \cite{31}. The output of two networks are combined and then put to the subsequent layers of the A3C's network. Once the agent is trained to understand the goal map, we can modify the target in the goal map to change the behaviors of the agent in real time without conducting a training process.

\begin{figure}[!t]
\centering
\includegraphics[width=0.5\linewidth]{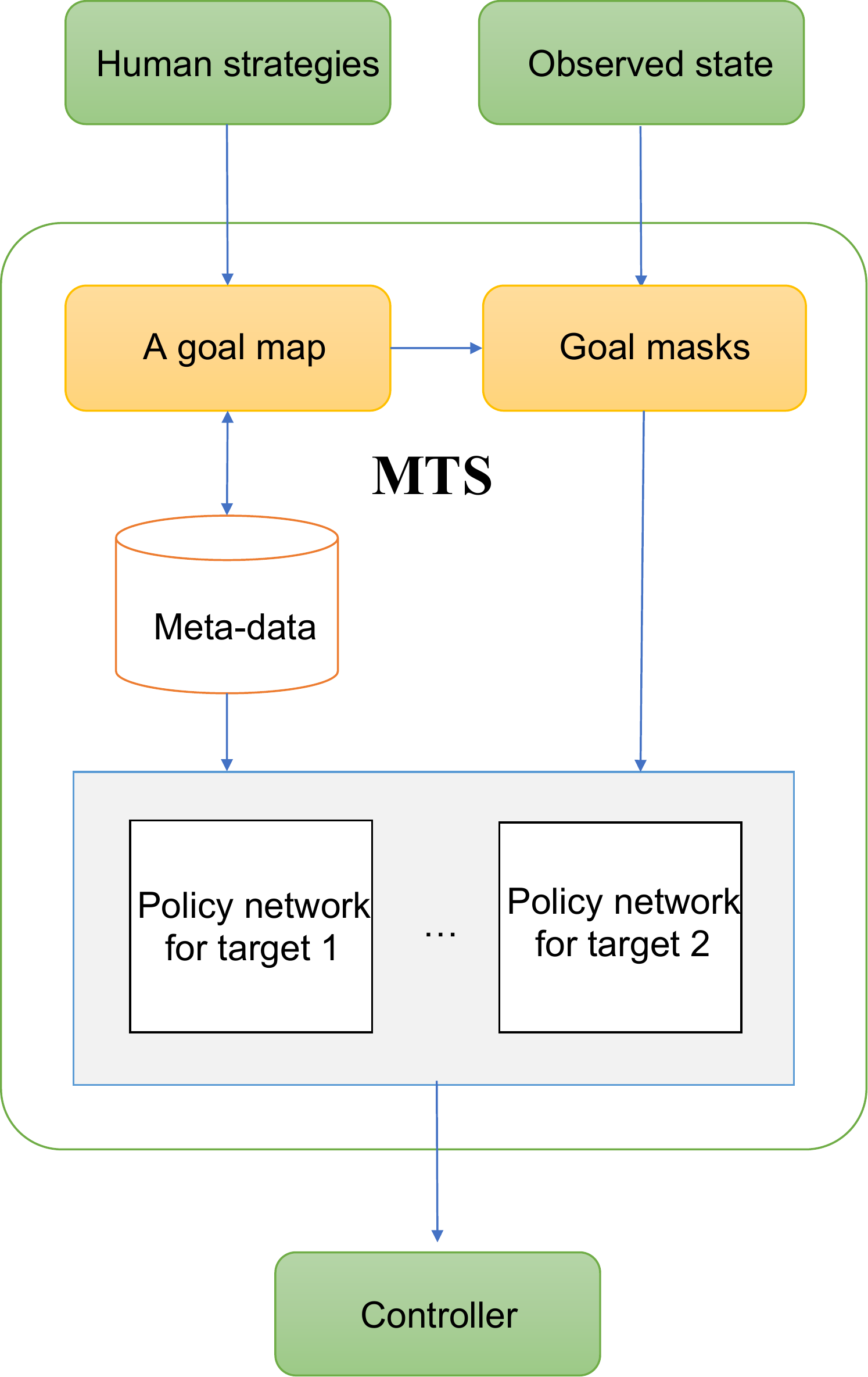}
\caption{A high-level system architecture of MTS.}
\label{fig:4} 
\end{figure}

\subsection{Multi-Target System}
\label{sec:3.4}
We put the concept of a goal map into a high-level system architecture, i.e. MTS, as shown in Fig.~\ref{fig:4}. In particular, MTS retrieves human strategies and creates a corresponding goal map. The goal map is used with the observed state to generate the goal masks. Reward signals, target locations, and target boundary regions are stored in a metadata structure. These data are used together with the goal masks to train different policy networks. Each policy network is used to train a set of agents which have the same targets. 

MTS can apply in a variety number of applications, especially in heterogeneous multi-agent systems. For each type of agents, we create a separate policy network. The MTS then schedules the agents' activities to cooperatively accomplish a designated task.

\section{Performance Evaluation}
\label{sec:4}

\subsection{Experimental Settings}

\begin{figure*}[!h]
\centering
\includegraphics[width=0.98\linewidth]{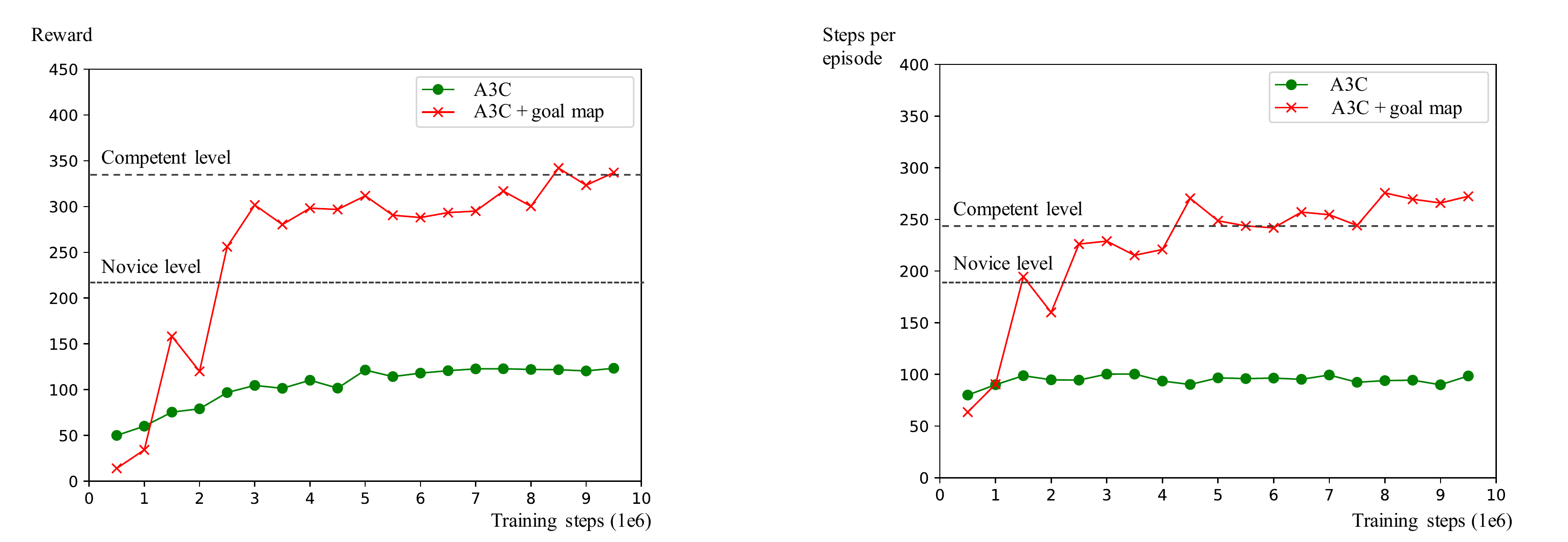}
\caption{A comparison of mean total of reward and mean total of steps per episode between two schemes: A3C and A3C with the goal map.}
\label{fig:6} 
\end{figure*}

In this section, we present the simulation by using Multiple Tank Defence to evaluate the proposed scheme in two-player mode. As explained earlier, we use A3C as the baseline method. The A3C's network parameters are kept the same as in \cite{10}, except the following changes. Each A3C algorithm is ran in a 8-core CPU. The learning rate is 0.004. The convolutional network includes two layers: one layer has 16 filters of size 8 $\times$ 8 with a stride of 4 and another layer has 32 filters of size 4 $\times$ 4 with a stride of 2. Finally, the action repeat of 5 is used while processing input.

Each A3C variant is trained in 10 million steps. This takes 2 training days for A3C baseline and 3 training days for A3C with a goal map. We evaluate the policy network in 20 different milestones during the training process. In each milestone, we measure the mean total of reward and the mean total number of steps per episode by running policy network in 50,000 steps. Finally, we include two human reference levels: novice level and competent level. The novice level is measured by letting a human plays the game in 50,000 steps and the competent level is measured by playing the game up to 500,000 steps.

\subsection{A3C with Goal Map}

We compare the performance of two schemes: A3C and A3C with the goal map. Fig.~\ref{fig:6} shows that the A3C method with the goal map performs 200\% better than the baseline. It also surpasses the human competent level. In the A3C baseline, two players do not cooperate as they greedily destroy enemies rather than protecting the base. In contrast, the use of a goal map aids two agents to cooperate to protect the base. This strategy prolongs the base's lifetime and hence obtains a high score in the long run.

Finally, we use MTS to modify agent behaviors and derive the following methods:

\begin{itemize}

\item In the single-player mode, we change the targets of the yellow region in real time based on current situation and human experience.

\item In the multi-player mode, we train the yellow agent using the goal map and a human plays in a role of the green player.

\item We modify the metadata structure to narrow the working region of two players so that the yellow player only focuses shooting on the right side and the green player only focuses shooting on the left side. This scheme achieves the highest score among three variants. Table \ref{table} summarizes the mean total of reward and mean total of steps per episode among three variants mentioned here. 
\end{itemize}

\begin{table}[h]
  \caption{Experimental results on different methods using goal map and MTS.}
  \label{table}
  \centering
  \begin{tabular}{llll}
    \toprule
    Scheme name & Mean reward & Mean steps/episode\\
    \toprule
    A3C with a    & 149 & 152\\
    goal map (single-&&\\
    player mode)&&\\
    \midrule
    A3C with a &301&234\\
    goal map&&\\
    and a human&&\\
    \midrule
    A3C with a&363&295\\
    goal map + &&\\
    modifying&&\\
    behaviors&&\\
    \bottomrule
  \end{tabular}
\end{table}

\section{Conclusions}
\label{sec:5}

This paper proposed the novel concept of the goal map and the high-level system architecture MTS. By using MTS, agents' behaviors can be modified in real time without conducting a training process. Moreover, agents can achieve an expected solution without using a burden of human feedback. As a result, agents do not become stuck in local minimum solutions and act more human-like in biased environments such as Multiple Tank Defence. Finally, the use of the goal map and MTS eases the cooperation between agents--agents or humans--agents in heterogeneous systems. Therefore, the study provides a promising framework that aims to attract considerable attention to building a human-like agent in large-scale systems with deep RL.






\end{document}